\documentclass{vgtc}                          

\ifpdf
  \pdfoutput=1\relax                   
  \usepackage{graphicx}                
  \DeclareGraphicsExtensions{.pdf,.png,.jpg,.jpeg,.svg} 
\else
  \usepackage{graphicx}                
  \DeclareGraphicsExtensions{.eps}     
\fi%

\graphicspath{{figures/}{pictures/}{images/}{./}} 

\usepackage{microtype}                 
\PassOptionsToPackage{warn}{textcomp}  
\usepackage{textcomp}                  
\usepackage{mathptmx}                  
\usepackage{times}                     
\usepackage{cite}                      
\usepackage{tabu}                      
\usepackage{booktabs}                  
\usepackage{enumitem}

\usepackage{authblk}
\usepackage{amsfonts}
\usepackage{mathtools}
\usepackage{color}

\usepackage{svg}

\renewcommand{\epsilon}{\varepsilon}

\onlineid{1001}

\vgtccategory{Research}

\vgtcinsertpkg

\title{Towards Large-Scale Rendering of Simulated Crops for Synthetic Ground Truth Generation on Modular Supercomputers}

\author[1,2]{Dirk Norbert Helmrich \thanks{E-mail: d.helmrich@fz-juelich.de\vspace{-12pt}}}
\author[1]{Jens Henrik G\"obbert}
\author[1]{Mona Giraud}
\author[1]{Hanno Scharr}
\author[1]{Andrea Schnepf}
\author[1,2]{Morris Riedel}
\affil[1]{Forschungszentrum J\"ulich GmbH, Germany}
\affil[2]{University of Iceland, Reykjav\'{i}k, Iceland}

\teaser{
\vspace{-1em}
\includegraphics[width=\linewidth]{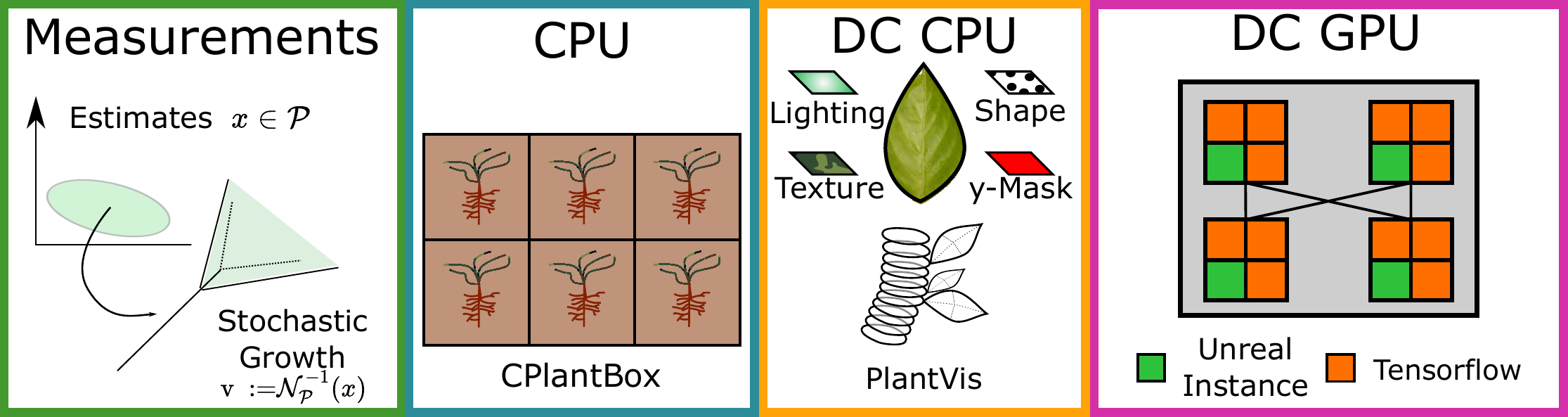}
\caption{\label{fig:datasets}Illustration of data and rendering pipeline. We are distributing CPlantbox over normal CPU nodes, with dynamics equations solved on the cores. Geometrization and texture generation requires more memory, making the task more optimal for larger data-centric (DC) CPU nodes. We can use our GPU nodes for Unreal instances that directly report data to TensorFlow.}
}

\abstract{Computer Vision problems deal with the semantic extraction of information from camera images.
Especially for field crop images, the underlying problems are hard to label and even harder to learn, and the availability of high-quality training data is low.
Deep neural networks do a good job of extracting the necessary models from training examples.
However, they rely on an abundance of training data that is not feasible to generate or label by expert annotation.
To address this challenge, we make use of the Unreal Engine to render large and complex virtual scenes.
We rely on the performance of individual nodes by distributing plant simulations across nodes and both generate scenes as well as train neural networks on GPUs, restricting node communication to parallel learning.
} 

\CCScatlist{
  \CCScatTwelve{Human-centered computing}{Visu\-al\-iza\-tion}{Visualization design and evaluation methods}{};
  \CCScatTwelve{Artificial Intelligence}{Computer Vision}{Computer Vision Problems}{Image Segmentation}
}

\begin{document}


\firstsection{\label{sec:int}Introduction}

\maketitle

\begin{figure}[h]
\includegraphics[width=0.5\linewidth]{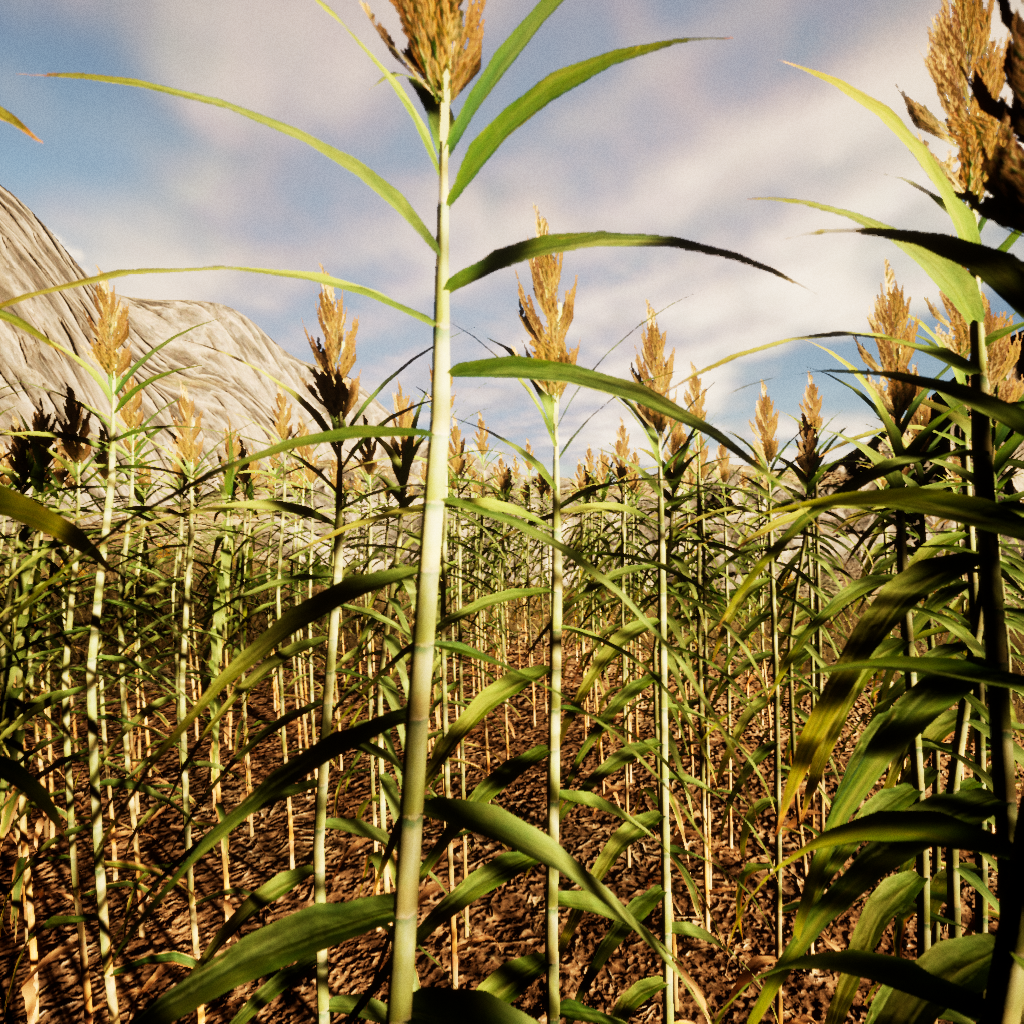}
\hspace{-3pt}
\includegraphics[width=0.5\linewidth]{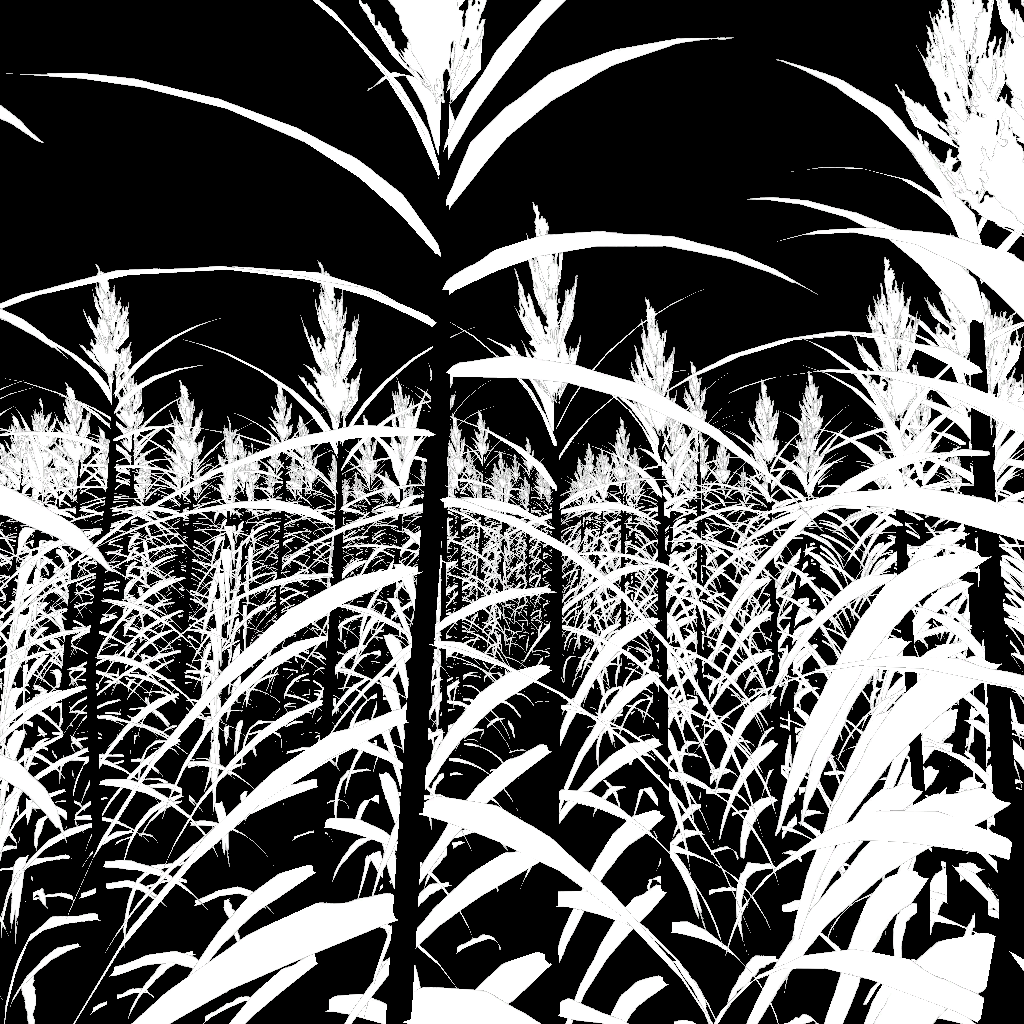}
\\
\includegraphics[width=0.5\linewidth]{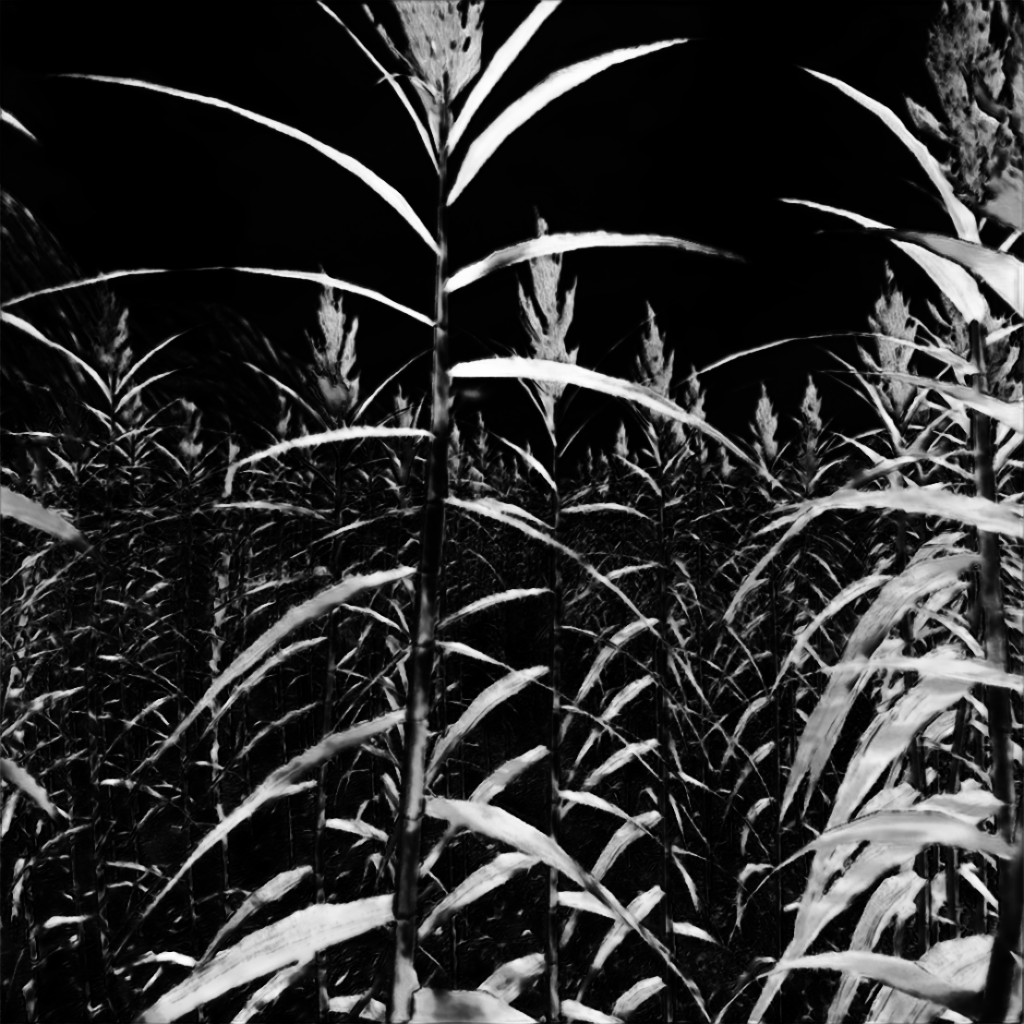}
\hspace{-3pt}
\includegraphics[width=0.5\linewidth]{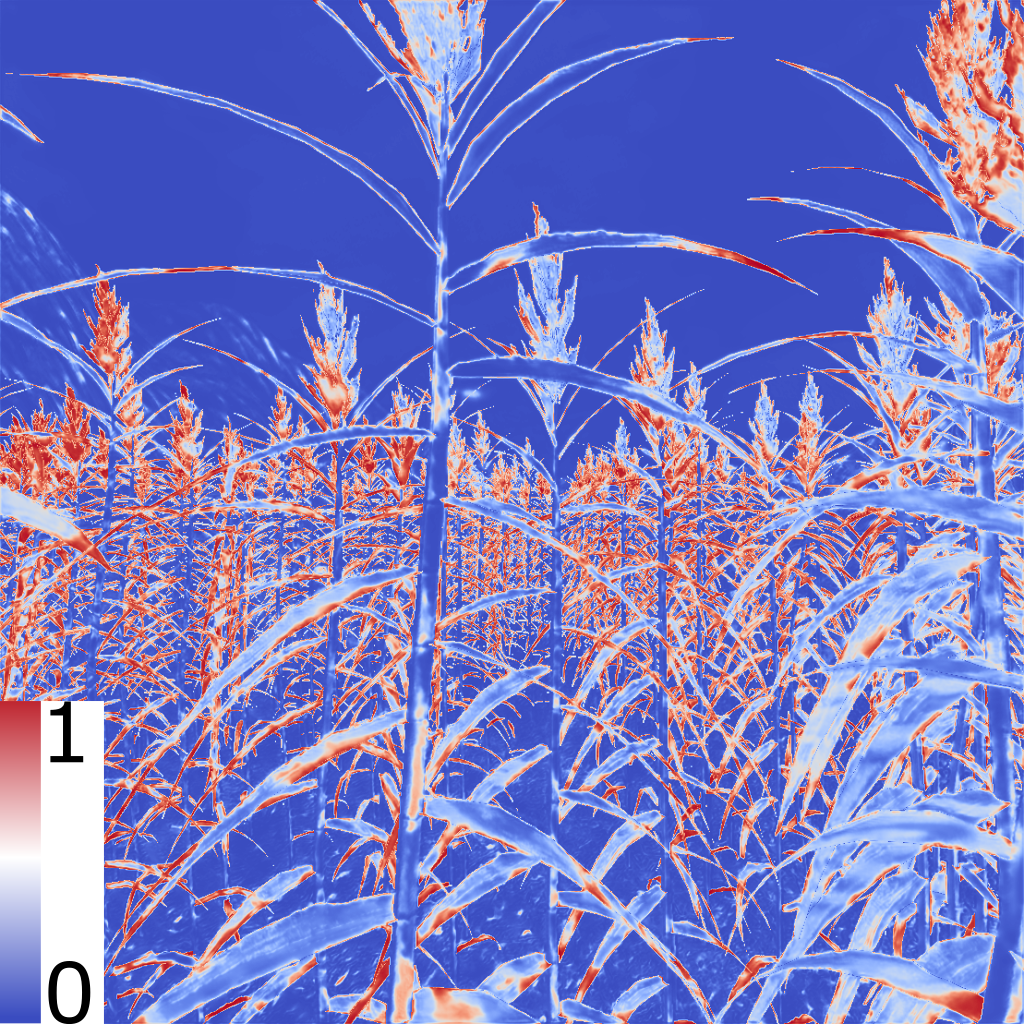}
\caption{\label{fig:syn}Preliminary testing of U-NET model \cite{unet} on a virtual scene with sample geometry \cite{reed}. Top left: Unreal rendering, including lense effects and strong shadows. Top right: ground truth extracted from virtual scene. Bottom left: Segmentation map produced by U-NET. Bottom right: Difference in segmentation to ground truth. The U-Net has most problems with finding sharp borders of objects and it produces a lower likelihood estimate for background plants.}
\vspace{-6pt}
\end{figure}

This poster focuses on supporting crop researchers with automated algorithms.
Computer vision problems factor into these algorithms very significantly, ranging from problems like segmentation to reconstruction.
This use-case has some unique difficulties, since field-level camera footage contains an extremely large amount of distinct objects and features.
To accommodate the great need for high quality training data, we want to use plant simulations, which facilitate the generation of ground truth data.
The combination of field measurements and simulation models is generally very promising \cite{phenoprom}.
One of the main issues with training artificial neural networks on crop images is that generating training data is a tedious process \cite{phenodeep}.
There are published use-cases in which synthetic data was used to train learning models, such as the UnrealPerson workflow \cite{unrealperson} that incorporates virtual humans for re-identification.
Synthetic data can help with this issue for several reasons:
\begin{itemize}[noitemsep]
\item Presence of ground truth
\item No ambiguity of local semantic assignment
\item Data augmentation is much more powerful
\item Parameter-space sampling and procedural enhancement \cite{deepsyn} can improve training data diversity
\end{itemize}
We aim to connect CPlantBox, a plant simulation tool that produces functional plant models \cite{cplant} and the Unreal Engine \cite{ue} to produce synthetic ground truth data at an extremely large scale on our modular supercomputer.
In the presented use-case of field crop images, algorithms have a hard time uncovering hidden mechanics even from very good label sets.
The reason for this is that these images suffer from weather conditions, lighting conditions, shadows, and background vegetation.
It can be very difficult for a machine learning algorithm to deal with these factors, even with full ground truth available.
This is illustrated in Figure \ref{fig:syn}, showing mislabeled image regions by a synthetically learned U-Net \cite{unet} on a rendered reed image.
The background of the image becomes noisy very quickly as more individual plants congregate.
This makes shape and feature detection more difficult and the algorithm will have higher error rates at greater distance to the camera.

\vspace{-0.5em}
\section{\label{sec:acc}Making use of modularity}

We are working with a modular supercomputer \cite{jureca}, consisting of different modules, from general-purpose computing to data-intensive accelerated computing. 
The planned pipeline has very heterogeneous components that require data-intensive computations.
This fact, however, also enables us to make better use of cluster modularity by specializing nodes for certain jobs.
Figure \ref{fig:datasets} shows certain steps in our pipeline distributed across nodes.
There are some important factors to consider for a successful implementation of this.

CPlantBox uses tropism parameters and plant parametrization.
It simulates growth in a node-tree-like structure \cite{cplant}.
This structure is especially important, as it does not only play an important role in plant geometry but also in subsequent plant- or field-level simulations of water and nutrient dynamics.
The simulation uses inherently stochastic parameters.
This way, simulations distributed across nodes will generate varying structures.
Parallelization of dynamics simulations can be done within a node, allowing for more expedient parallelization of difficult dynamics problems.

We aim to employ a robust geometrization and animation pipeline to facilitate the photorealistic rendering of these plant geometries.
In addition to basic geometrization, we are planning to implement a mixture approach of stem skeletonization and leaf movement simulation.
To facilitate this, the plant geometries have to be preconditioned for this use in the rendering pipeline.

Moving to the Unreal Engine, a single instance is run on an accelerated node.
The advantage of this is that scene configurations can be distributed across nodes.
Scene-specific properties, such as sun position, weather, and wind are not only great for distribution, but also improve robustness of the training minibatch in case of gradient averaging.
Every accelerated node has access to its own subset of data.
However, while gradient averaging is one answer to handling the training process, we want to investigate different distribution methods that could benefit from synthetic data generation.
With a minibatch that is so diverse regarding image construction, the immediate question is whether the machine learning model has enough capacity to learn the problem posed by the synthetic data generation.
As seen in Figure \ref{fig:syn}, background identification of smaller plants that still belong to the scene becomes progressively harder.
The next challenge would be identifying whether specific scene configurations would improve results, or if the model capacity is reached.

\vspace{-0.5em}
\section{\label{sec:err}Online Learning}

\begin{figure}[h]
\includegraphics[width=\linewidth]{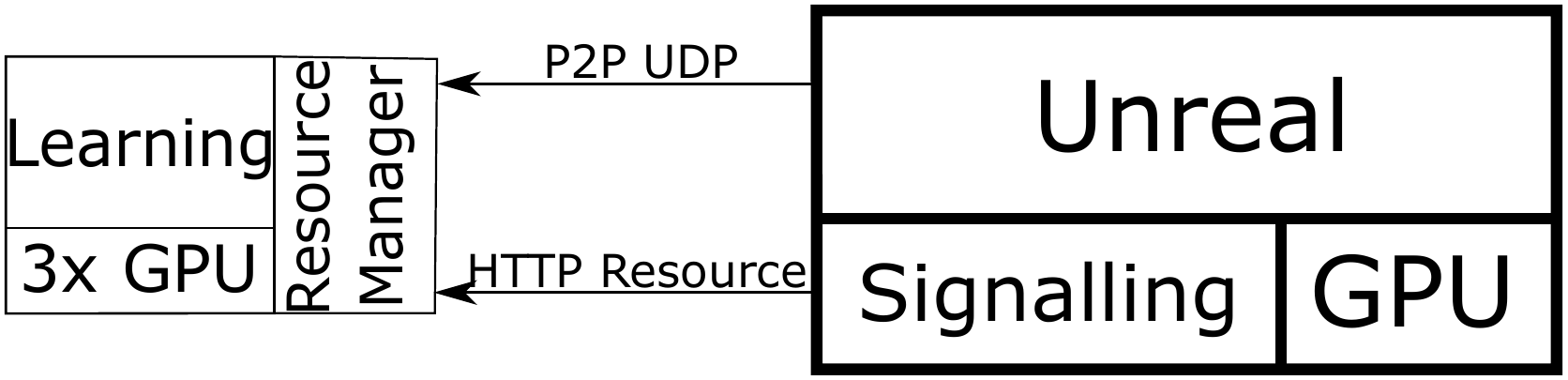}
\caption{\label{fig:rtc}Illustration of Unreal Engine setup on an individual node. The signalling server is a lightweight node.js application handling HTTP resources and brokering of the WebRTC connection.}
\end{figure}

With the introduction of the 4.27 version of the Unreal Engine, Epic Games will introduce a production-ready PixelStreaming addon that utilizes WebRTC \cite{rtc}.
The Unreal Engine packages frames on the graphics card and sends them via the connection brokered by the signaling server.
To limit node communication, we are collecting this stream of images on the same node as Unreal.
This way, we can employ distributed machine learning and use the interconnect exclusively for model training.
The focus of optimization in this context  is less about interactivity than it is about predictability of performance and workload.
A stable rendering frame rate is more beneficial than aiming for a high frame rate, since this enables an easier utilization of the pipeline as a whole.
The primary reason to use a separate server instead of directly connecting Unreal with the AI Framework is integratability.
This setup allows the use of the standard git version of Unreal Engine and utilizes it to serve our framework as opposed to using a tailored version of a large library.
Figure \ref{fig:rtc} illustrates our individual node setup.

\vspace{-0.5em}
\section{\label{sec:concl}Conclusion}

We are confident in our approach and will move forward with the realization of the pipeline.
We would like to discuss infrastructural and technical aspects of this.
We would also welcome feedback on future uses, simulation data integration, as well as visualization of learning progress in a setting in which data is infinite.

\acknowledgments{
The authors would like to acknowledge funding provided by the German government to the Gauss Centre for Supercomputing via the InHPC-DE project (01|H17001).
}

\bibliographystyle{abbrv-doi}

\bibliography{phddirk}
\end{document}